\documentclass[12pt]{l4dc2023}

\usepackage{cases}
\usepackage{amsfonts}
\usepackage{mathtools}
\usepackage{mathrsfs}
\usepackage{nicefrac}
\usepackage{cite}
\usepackage{multirow}
\usepackage{bm}
\usepackage{soul}
\setul{}{0.8pt}
\usepackage[font=footnotesize,labelfont=bf]{caption}

\newcommand{\R}{\mathbb{R}}
\newcommand{\Xsafe}{\mathcal{X}_{\text{safe}}}
\newcommand{\Xunsafe}{\mathcal{X}_{\text{unsafe}}}
\newcommand{\xsafe}{x_{\text{safe}}}
\newcommand{\xunsafe}{x_{\text{unsafe}}}
\newcommand{\usafe}{u_{\text{safe}}}

\newcommand{\norm}[1]{\left\lVert #1\right\rVert}
\DeclareMathOperator*{\argmin}{arg\,min}

\title[In-Distribution Barrier Functions]{In-Distribution Barrier Functions: Self-Supervised Policy Filters that Avoid Out-of-Distribution States}

\usepackage{times}

\author{%
 \Name{Fernando Casta\~neda} \Email{fcastaneda@berkeley.edu}\\
 \addr University of California, Berkeley
 \AND
 \Name{Haruki Nishimura} \Email{haruki.nishimura@tri.global}\\
 \addr Toyota Research Institute, Los Altos, CA%
 \AND
 \Name{Rowan McAllister} \Email{rowan.mcallister@tri.global}\\
 \addr Toyota Research Institute, Los Altos, CA%
 \AND
 \Name{Koushil Sreenath} \Email{koushils@berkeley.edu}\\
 \addr University of California, Berkeley
 \AND
 \Name{Adrien Gaidon} \Email{adrien.gaidon@tri.global}\\
 \addr Toyota Research Institute, Los Altos, CA%
}

\begin{document}

\maketitle

\vspace{-1.5em}

\begin{abstract}%
Learning-based control approaches have shown great promise in performing complex tasks directly from high-dimensional perception data for real robotic systems. 
Nonetheless, the learned controllers can behave unexpectedly if the trajectories of the system divert
from the training data distribution,
which can compromise safety.
In this work, we propose a control filter that wraps any reference policy and effectively encourages the system to stay in-distribution with respect to offline-collected safe demonstrations.
Our methodology is inspired by Control Barrier Functions (CBFs), which are model-based tools from the nonlinear control literature that can be used to construct minimally invasive safe policy filters.
While existing methods based on CBFs require a known low-dimensional state representation, our proposed approach is directly applicable to systems that rely solely on high-dimensional visual observations by learning in a latent state-space.
We demonstrate that our method is effective for two different visuomotor control tasks in simulation environments, including both top-down and egocentric view settings.

\end{abstract}


\section{Introduction}
\label{sec:01introduction}

The modern advances in the representation learning literature have been an enabling factor for the recent surge of a wide variety of methods for robotic control directly from images or high-dimensional sensory observations \citep{watter2015embed, ebert2018visual, hafner2019learning, lenz2015deepmpc, zhang2019solar, van2016stable}. These approaches for visuomotor planning and control have the potential to solve challenging tasks in which the state of the system might not be directly observable, or even not possible to model analytically. While promising, the high-dimensionality of the problem make these methods susceptible to several open challenges.
For example, the exploration requirements of reinforcement learning (RL) algorithms are significantly exacerbated for these tasks, due to the high-dimensionality of the observations. This means that trying to learn safe control policies using RL often requires us to accept that abundant failures will occur during training. On the other hand, supervised learning approaches for control, such as behavioral cloning, would in principle seem less prone to exhibit unsafe behaviors. However, it is well known that simply because of their data-dependent nature, these methods are still susceptible to a key challenge named \emph{distributional shift}: if the trajectories of the system divert from the training data distribution, the controller might take unexpected actions.

On the other hand, the control theory literature extensively covers the problem of long-horizon constraint satisfaction. In particular, Control Barrier Functions (CBFs, \citet{ames2014cbf}) are a popular model-based tool used to restrict the trajectories of the system from entering undesirable regions of the state-space. One of the properties of CBFs that explain their recent popularity is that they decouple the problem of constraint-satisfaction from any performance objective. Specifically, if a CBF is available, then \citet{ames2014cbf} showed that 
we can construct a minimally invasive safety filter that transforms into safety-preserving control actions any unsafe commands that an arbitrary reference policy could output.

The main question we want to address in this work naturally emerges from the previous discussion: can we take inspiration from CBFs to avoid out-of-distribution (OOD) states when using data-driven controllers for visuomotor tasks? Even though CBFs are model-based tools that, as such, require knowledge of the state-space and dynamics of the system, the recent advances on learning latent state-space representations and associated dynamics models clearly set a path for linking data-driven visuomotor policy learning with the use of model-based control-theoretic tools such as CBFs.

\textbf{Contributions}: We present an end-to-end self-supervised approach for learning a task-agnostic policy filter which prevents the system from entering OOD states. We do not assume knowledge of the state-space or system dynamics. In addition, our framework only requires an offline-collected dataset
of safe demonstrations (where the concept of safety is only linked to the demonstrator's subjectivity, as it is their responsibility to provide the dataset). We therefore do not require any unsafe demonstrations to learn a safe policy filter, in contrast to most other works tackling constrained policy learning. Furthermore, to the best of our knowledge, this is the first work that uses CBFs for constructing policy filters in learned latent state-spaces. This endows our approach with the flexibility of being applicable to systems with high-dimensional sensory observations, in contrast to most prior CBF-based methods. We present simulation experiments on two different visuomotor control tasks, which suggest that our framework, taking only raw RGB images as input, can learn to significantly reduce the distributional shift from safe demonstrations and, consequently, critically improve the safety of both systems.

\section{Related Work}

There exist some constructive procedures for synthesizing CBFs based on sum-of-squares programming \citep{jarvis2003some, majumdar2013control, dai2022convex, wang2022safety} or Hamilton-Jacobi reachability \citep{choi2021robust}. However, these methods require knowledge of the dynamics of the system and typically suffer from scalability issues for high-dimensional systems. More in line with this work, some recent results show that CBFs can be learned from data \citep{jin2020neural, dawson2022safe, qin2022sablas, robey2020learning, lindemann2021learning, abate2021fossil, jagtap2020control}. None of these works, however, consider systems with high-dimensional observations. Furthermore, the works \citet{jin2020neural, dawson2022safe, qin2022sablas} assume a priori knowledge of a control-invariant safe set, and focus on building a CBF for that particular set. The line of research of \citet{robey2020learning, lindemann2021learning} has the most similar problem setup to our work, as they also consider learning from safe demonstrations. Although, notably, the authors provide formal verification arguments for the learned CBFs, their methods are not applicable to high-dimensional observations, assume a nominal dynamics model is given, and use an algorithmic approach for the detection of the boundary of the dataset that does not scale to large datasets. Other recent approaches build signed distance functions from sensory measurements that are obtained from a LiDAR or stereo cameras \citep{long2021learning, srinivasan2020synthesis, cosner2022self}. However, these functions are not encouraged to satisfy any set invariance property.

Extensions of the CBF-based control filters to systems with dynamics or measurement-model uncertainty have also been recently proposed \citep{nguyen2021robust, castaneda2021pointwise, taylor2020towards, dhiman2021control, dean2021guaranteeing}. These works assume that a CBF is provided, and formulate uncertainty-robust optimization problems for the controller design. They can be considered complementary to our deterministic but end-to-end approach. Future work should explore quantifying uncertainty estimates within our framework to robustify the learned policy filters.

Several existing approaches for OOD prevention learn density models of training data that can be then used to restrict the agent from taking low likelihood actions or moving towards unvisited states \citep{mcallister2019robustness, richter2017safe, wu2019behavior, kumar2019stabilizing}. Although some of these methods have been shown to be effective at offline RL settings that are specially susceptible to distributional shift, the learned density models have no notion of control invariance and, therefore, do not consider the problem of how to prevent distributional shift over a long time horizon. A notable exception is the work of \citet{kang2022lyapunov} to constrain long-term distributional shift, in which a min-max Bellman backup operator is constructed so that Lyapunov-like functions arise as value functions of an offline RL problem. This work however does not consider the extension to visuomotor control tasks in learned latent spaces. Furthermore, for our approach we choose not to rely on a min-max backup operator to learn the certificate function and,
instead, use the very suitable theory of CBFs to devise a self-supervised learning framework.

Finally, the work of \citet{wilcox2022ls3} presents a framework to learn safe sets in a latent state-space for iterative control tasks. Compared to this work, our framework has the advantage that it is task-agnostic and does not require any interactions with the environment during training.

\section{Background on Control Barrier Functions}
\label{sec:02background}
We start by introducing some necessary background on Control Barrier Functions, which are tools from the nonlinear control literature that serve to enforce safety constraints for systems with known dynamics. As will be clear later, CBFs are particularly well-suited for continuous-time nonlinear control-affine systems of the form
\vspace{-0.5em}
\begin{equation}\label{eq:system}
    \dot{x} = f(x) + g(x)u,
\end{equation}
where $x \in \mathcal{X} \subset \R^n$ is the state and $u \in \mathcal{U} \subset \mathbb{R}^m$ the control input. We assume that $f: \mathcal{X} \to \R^n$ and $g: \mathcal{X} \to \R^{n\times m}$ are locally Lipschitz continuous.

In the CBF literature, safety is considered as a set invariance problem. In particular, we say that a control policy $\pi: \mathcal{X} \to \mathcal{U}$ assures the safety of system \eqref{eq:system} with respect to a set $\Xsafe \subset \mathcal{X}$ if the set $\Xsafe$ is forward invariant under the control law $\pi$, i.e., for any $x_0 \in \Xsafe$, the solution $x(t)$ of system \eqref{eq:system} under the control law $\pi$ remains within $\Xsafe$ for all $t \geq 0$.

\begin{definition}[Control Barrier Function, \citet{ames2017cbf}]
\label{def:cbf}
We say that a continuously differentiable function $B: \mathcal{X} \to \R$ is a \emph{Control Barrier Function (CBF)} for system \eqref{eq:system} with associated safe-set $\Xsafe \subset \mathcal{X}$ if the following three conditions are satisfied:
\vspace{-0.5em}
\begin{align}
    B(x) \geq 0\ \ \  &\forall x \in \Xsafe, \label{eq:cbf-def-cond-safe}\\
    B(x) < 0\ \ \  &\forall x \in \mathcal{X}\setminus \Xsafe, \label{eq:cbf-def-cond-unsafe}\\
    \exists u \in \mathcal{U}\ \ \text{s.t.}\  \ \dot{B}(x,u)+ \gamma (B(x)) \geq 0\  \ \ &\forall x \in \mathcal{X}, \label{eq:cbf-def-cond-ascent}
\end{align}
\vspace{-0.5em}
where $\gamma: \R \to \R$ is an extended class $\mathcal{K}_\infty$ function.
\end{definition}

The existence of a CBF $B$ guarantees that for system \eqref{eq:system} any Lipschitz continuous control policy $\pi$ satisfying
\vspace{-0.5em}
\begin{equation}
    \label{eq:cbf_condition}
    \pi(x) \in \{u\in \mathcal{U} : \underbrace{\nabla B (x) [f(x) + g(x) u]}_{=\dot{B}(x,u)} + \gamma\left(B(x) \right) \geq 0\}
\end{equation}
will render the set $\Xsafe$ forward invariant \citep[Corollary~2]{ames2017cbf}.

For a given task-specific reference controller $\pi_{\text{ref}}: \mathcal{X} \to \mathcal{U}$ that might be safety-agnostic, the condition of \eqref{eq:cbf_condition} can be used to formulate an optimization problem that, when solved at every time-step, yields a minimally-invasive policy safety filter \citep{ames2014cbf}:
\vspace{-0.5em}
\begin{align}
\pi_{\text{CBF}}(x) & =\underset{u\in \mathcal{U}}{\argmin}  \quad \norm{u-\pi_{\text{ref}}(x)}^2 \tag{CBF-QP} \label{eq:cbf-qp}\\
& \text{s.t.}\ \  \nabla B (x) [f(x) + g(x) u] + \gamma (B(x)) \geq 0. \notag
\end{align}
Assuming that the actuation constraints that define $\mathcal{U}$ are linear in $u$, this problem is a quadratic program (QP). This is a consequence of the dynamics of the system \eqref{eq:system} being control-affine, and it practically means that the problem can be solved to a high precision very quickly (around $10^3$Hz). This is critical since the CBF-QP needs to be solved at the real-time control frequency.
\section{Problem Statement}
\label{sec:03problemstatement}
The CBF-QP constitutes a very appealing approach for practitioners: it provides a task-agnostic minimally invasive filter that can wrap safety around any given policy $\pi_{\text{ref}}$, and therefore rewrite any unsafe control input that $\pi_{\text{ref}}$ could output at any time. However, designing a valid CBF is nontrivial. In fact, it is still an active research topic even when assuming perfect knowledge of the dynamics of the system \citep{dai2022convex, choi2021robust, wang2022safety}. The two main difficulties in the design of a CBF are the following: first, a control-invariant set $\Xsafe$ must be obtained (which in general is different from the geometric constraint set that could be obtained, for instance, from a signed-distance field) and, second, a function that satisfies condition \eqref{eq:cbf-def-cond-ascent} must be found for that set. Furthermore, even after obtaining a CBF, solving the CBF-QP requires perfect state and dynamics knowledge.

With our framework, we take initial steps towards building a safe policy filter from high-dimensional observations.
Specifically, we take inspiration from CBFs to design an end-to-end learning framework to constrain deep learning models to remain in-distribution of the training data. We take as input a dataset of high-dimensional observations of different safe demonstrations, and build a neural CBF-like function that encourages the system to always stay in-distribution with respect to the observations from the safe demonstrations. This, in turn, significantly improves the safety of the system during deployment.

More concretely, for a given dataset of $N$ safe trajectories $\mathbb{D} = \left\{\left(I^i_t, u^i_t\right)^{t=T_i}_{t=0}\right\}^{i=N}_{i=1}$ we tackle the problem of designing a policy filter that can be applied to any reference controller $\pi_{\text{ref}}$ to detect and override actions from $\pi_{\text{ref}}$ that lead to OOD states. We denote $I^i_t$ and $u^i_t$ the high-dimensional observation and control input, respectively, measured at time $t$ for the $i$th trajectory. Furthermore, $T_i$ is the final time-step of trajectory $i$.
The demonstrations in the dataset $\mathbb{D}$ might correspond to different tasks and they do not need to be optimal with respect to any objective. In fact, our only assumption is that the dataset only contains safe demonstrations (in the sense that these trajectories should not contain any states from which the system is deemed to fail, even if it has not failed yet), so that we can encourage long-term constraint satisfaction using CBFs.
\section{In-Distribution Barrier Functions}
\label{sec:04learningcbfs}
In this section, we introduce a self-supervised approach for synthesizing neural CBF-like functions whose aim is to constrain the system to remain in-distribution with respect to an offline dataset of safe demonstrations. We call these functions \emph{in-Distribution Barrier Functions} (iDBFs).
We will for now assume that we have a parametric continuous-time control-affine model of the dynamics of the system in a state-space $\mathcal{X} \subset \R^n$
\vspace{-1em}
\begin{equation}\label{eq:neural-system}
    \dot{x} = f_\theta(x) + g_\theta(x)u,
    \vspace{-0.5em}
\end{equation}
and present the iDBF learning procedure for this system. Furthermore, for this section we assume that the dataset $\mathbb{D}$ of safe trajectories contains true state measurements, i.e., $\mathbb{D} = \left\{\left(x^i_t, u^i_t\right)^{t=T_i}_{t=0}\right\}^{i=N}_{i=1}$, where $x^i_t$ and $u^i_t$ are the state and control input, respectively, measured at time $t$ for the $i$th trajectory. In Section \ref{sec:05learningcbfsinlatentspace}, we will provide details on how to learn a dynamics model of this form in a latent state-space when we have a dataset containing high-dimensional sensory observations.

We parameterize an iDBF $B_\phi: \mathcal{X} \to \R$ as a neural network with parameters $\phi$, and construct an empirical loss function that encourages it to satisfy the three CBF conditions \eqref{eq:cbf-def-cond-safe}, \eqref{eq:cbf-def-cond-unsafe} and \eqref{eq:cbf-def-cond-ascent} with respect to a set $\Xsafe$ that is also implicitly learned through self-supervision. To design the loss function, we take inspiration from previous literature on learning CBFs \citep{dawson2022safe, qin2022sablas, chang2019neural}. Nevertheless, instead of assuming that the safe-set $\Xsafe$ is given and that we can sample from it and from its unsafe complement $\Xunsafe\doteq\mathcal{X}\setminus\Xsafe$, we build our loss function in a self-supervised manner just from the dataset of safe demonstrations. We accomplish this by leveraging ideas from contrastive learning \citep{gutmann2010noise, oord2018representation, chopra2005learning, weinberger2009distance, schroff2015facenet}. In particular, as we explain in detail later, we build a contrastive distribution from which to sample candidate unsafe states, given that we do not have any unsafe demonstrations in our dataset. The loss function we propose for learning an iDBF takes the following form:

\vspace{-1em}
{\small
\begin{multline}
    \label{eq:cbf-loss}
    \mathcal{L}_{\text{iDBF}} = \frac{w_{\text{safe}}}{N_{\text{safe}}} \sum_{\xsafe} \left[\epsilon_{\text{safe}} - B_\phi(\xsafe)\right]^{+} + \frac{w_{\text{unsafe}}}{N_{\text{unsafe}}} \sum_{\xunsafe} \left[\epsilon_{\text{unsafe}} + B_\phi(\xunsafe)\right]^{+} +\\
    \frac{w_{\text{ascent}}}{N_{\text{safe}}} \sum_{(\xsafe, \usafe)} \bigg[\epsilon_{\text{ascent}} -  \big(\nabla B_\phi (\xsafe) [f_\theta(\xsafe) + g_\theta(\xsafe) \usafe] + \gamma (B_\phi(\xsafe))\big)\bigg]^{+},
\end{multline}}
where $[\cdot]^+ \coloneqq \text{max}(0, \cdot)$; $(\xsafe, \usafe)$ are samples from the empirical distribution of the dataset $\mathbb{D}$; $\xunsafe$ are samples from a contrastive distribution that we will define soon; $w_{\text{safe}}$, $w_{\text{unsafe}}$ and $w_{\text{ascent}}$ are the weights of the different loss terms; and $\epsilon_{\text{safe}}$, $\epsilon_{\text{unsafe}}$ and $\epsilon_{\text{ascent}}$ are positive constants that serve to enforce strictly the inequalities and generalize outside of the training data.

The goal of the first two terms in the loss function is to learn an iDBF that has a positive value in states that belong to the data distribution of safe demonstrations, and negative everywhere else (meaning we are encouraging the satisfaction of conditions \eqref{eq:cbf-def-cond-safe} and \eqref{eq:cbf-def-cond-unsafe} of the definition of CBF). Note that this classification objective is very related to the notion of energy-based models (EBMs) ---neural network density models that assign a low energy value to points close to the training data distribution and a high value to points that are far from it \citep{hinton2002training}. In fact, we took inspiration from the Noise Contrastive Estimation (NCE, \citet{gutmann2010noise}) training procedure of EBMs, in particular the InfoNCE loss \citep{oord2018representation}, to design \eqref{eq:cbf-loss}. Intuitively, these methods use a noise contrastive distribution to generate candidate examples where to increase the value of the energy of the EBM, while decreasing the energy at the training data points.
We precisely want the opposite result for our problem: a high value of $B_\phi$ on the training data distribution, and a low value everywhere else. However, we have one additional requirement, which is that the iDBF should have a value of zero at the boundary, as set by conditions \eqref{eq:cbf-def-cond-safe} and \eqref{eq:cbf-def-cond-unsafe}. This is the reason why we design the two first terms of the loss function using the $[\cdot]^+$ operator.

In the third term of the loss \eqref{eq:cbf-loss}, note that we do not encourage the satisfaction of condition \eqref{eq:cbf-def-cond-ascent} over the entire state-space, but only over the dataset of safe demonstrations. However, in the definition of CBF, if condition \eqref{eq:cbf-def-cond-ascent} is only satisfied $\forall x \in \Xsafe$ instead of $\forall x \in \mathcal{X}$, the CBF still guarantees the control-invariance of $\Xsafe$.
We are therefore using our empirical data distribution of safe demonstrations as a sampling distribution covering the set $\Xsafe$, which we are also implicitly learning as the zero-superlevel set of $B_\phi$.
Furthermore, compared to prior approaches that encourage the satisfaction of condition \eqref{eq:cbf-def-cond-ascent} for a single policy \citep{dawson2022safe, qin2022sablas}, we instead use all pairs $(\xsafe, \usafe)$ present in the dataset $\mathbb{D}$ to compute this term of the loss. This way, we force the set of admissible control inputs \eqref{eq:cbf_condition} to be as large as our dataset allows, reducing the conservatism of the learned iDBF.

In order to generate the contrastive distribution from which to sample $\xunsafe$, as we ultimately want to learn the iDBF in a latent state-space in which it might not be intuitive how to construct a noise distribution, we take the following steps. 1) Based on the dataset of safe demonstrations $\mathbb{D}$, we train a neural behavioral cloning (BC) model that outputs a multi-modal Gaussian distribution over actions conditioned on the state, with density $\pi_{\text{BC}} (u | x)$. 2) Then, during the training process of the iDBF, for each $\xsafe$ state sampled from $\mathbb{D}$ we randomly take $N_{\text{candidate}}$ control inputs $u_{\text{candidate}}$ and evaluate their density value based on the BC model $\pi_{\text{BC}} (u_{\text{candidate}} | \xsafe)$. 3) If the value of the density falls below a threshold, then that control input is forward-propagated for one timestep using the dynamics model \eqref{eq:neural-system} to generate a sample $\xunsafe$. This way, we generate a contrastive data distribution by propagating actions that are unlikely present in the dataset of safe demonstrations. Furthermore, by only propagating these actions for one timestep, the contrastive distribution is close to the training data, which is desirable for the learning process \citep{gutmann2011bregman}.
\section{Learning iDBFs from High-Dimensional Observations}
\label{sec:05learningcbfsinlatentspace}

\begin{figure}
    \centering
    \vspace{-1.5em}
    \includegraphics[width=0.9\textwidth,trim={0 1.3cm 0 0},clip]{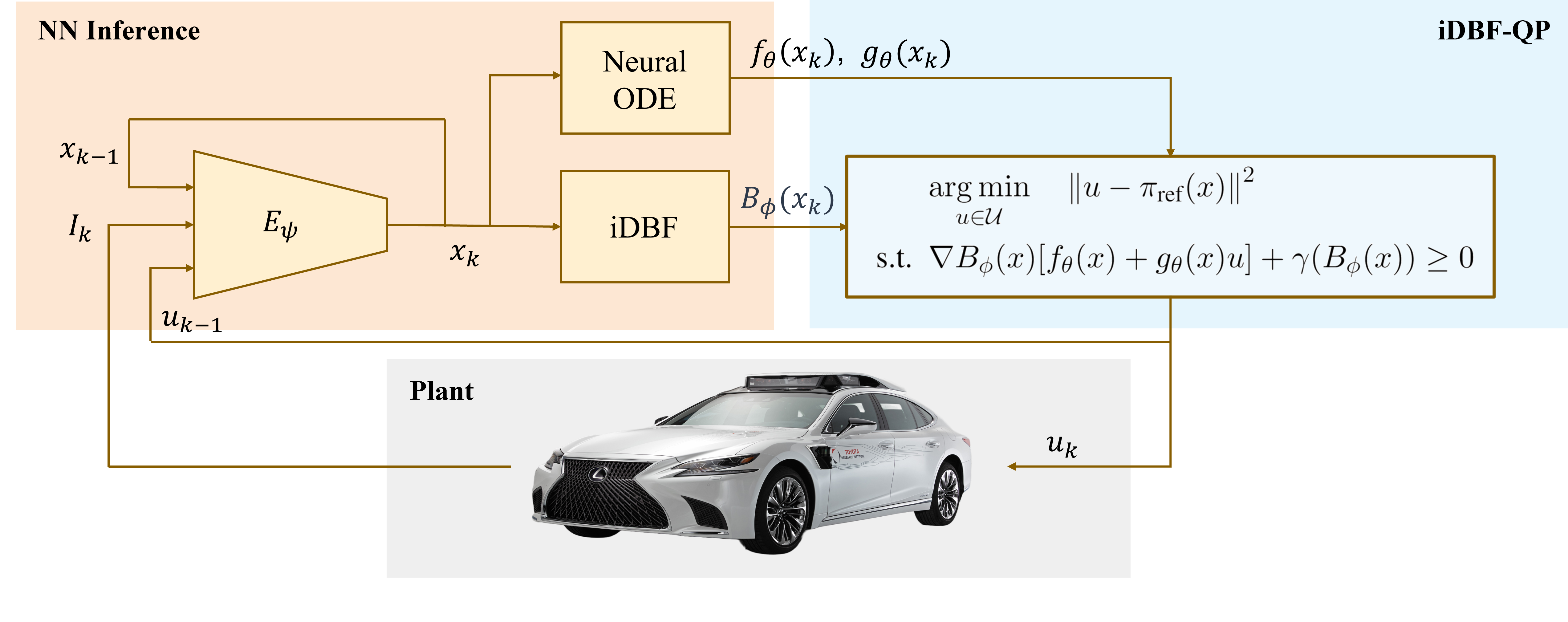}
    \caption{Our framework's inference diagram. At each timestep, based on the current observation $I_k$ and the previous latent state and action, the encoder network $E_\psi$ outputs a new latent state $x_k$. Then, the iDBF and dynamics networks give the values of $B_\phi(x_k)$, $f_\theta(x_k)$ and $g_\theta(x_k)$ which are passed to the iDBF-QP policy filter. The iDBF-QP takes a reference control input for the current timestep $\pi_{\text{ref}}(x_k)$ and returns the closest action that keeps the system in-distribution with respect to the offline-collected dataset of safe demonstrations. For both of the examples of Section \ref{sec:06examples}, the total inference time (NN Inference + solving the iDBF-QP) of our framework is less than $5$ milliseconds. }
    \vspace{-1em}
    \label{fig:diagram}
\end{figure}

After introducing the training procedure for an iDBF when the state representation and dynamics model \eqref{eq:neural-system} are given, we now relax these assumptions and present an approach to learn a latent state-space representation and a continuous-time dynamics model of the form \eqref{eq:neural-system}, suitable to be integrated in the same end-to-end learning framework. We therefore now consider precisely the problem setting described in Section \ref{sec:03problemstatement}, in which we only assume having access to a dataset containing observation-action pairs of safe demonstrations $\mathbb{D} = \left\{\left(I^i_t, u^i_t\right)^{t=T_i}_{t=0}\right\}^{i=N}_{i=1}$.

We use an autoencoder architecture to obtain the latent state-space representation, and employ the training procedure of Neural Ordinary Differential Equations (Neural ODEs, \citet{chen2018neural}) to learn a dynamics model of the form \eqref{eq:neural-system} in the latent state-space. Note that by enforcing the continuous-time control-affine structure of the dynamics model, we ensure that the iDBF-QP policy filter (equivalent to the CBF-QP, see Figure \ref{fig:diagram}) obtained with the learned iDBF and dynamics model will also be a quadratic program.

The inference procedure of our end-to-end learning framework is depicted in Figure \ref{fig:diagram}. We use a recursive encoder network $E_\psi$ that takes the current measurement $I_{k}$, as well as the previous latent state $x_{k -1}$ and action $u_{k -1}$ to generate the new latent state $x_{k}$ at each time-step $k$. The decoder network $D_\xi$ generates a reconstructed observation $\hat{I}_{k}$ for each latent state $x_{k}$. The proposed loss function for the latent state-space representation and dynamics model penalizes both the prediction error of the dynamics model and the observation reconstruction error:

\vspace{-1.4em}
{\scriptsize
\begin{multline}
    \label{eq:dynamics-loss}
    \hspace{-10pt}
    \mathcal{L}_{\text{dyn}} = \frac{1}{N_{\text{dyn}}(T_{\text{pred}}+1)} \sum_{j=1}^{N_{\text{dyn}}} \sum_{k=0}^{T_{\text{pred}}} \bigg[ w_{\text{state}} \norm{\tilde{x}_{t_j+k|t_j} - x_{t_j + k}}^2 + 
    w_{\text{rec}_1} \norm{\tilde{I}_{t_j+k|t_j} - I_{t_j + k}}^2 + w_{\text{rec}_2} \norm{\hat{I}_{t_j+k} - I_{t_j + k}}^2 \bigg].
    \vspace{-1em}
\end{multline}}
Here, $x_{t_j + k}= E_\psi( I_{t_j + k}, x_{t_j + k-1}, u_{t_j + k-1})$ is the latent state at timestep $t_j + k$. $\tilde{x}_{t_j+k|t_j}$ denotes the latent state prediction obtained by forward-propagating the dynamics model \eqref{eq:neural-system} to timestep $t_j + k$ starting from the state $x_{t_j}$ and using zero-order hold on the sequence of control inputs $(u_{t_j}, u_{t_j + 1}, ..., u_{t_j + k -1})$. Additionally,
$\tilde{I}_{t_j+k|t_j} \coloneqq D_\xi(\tilde{x}_{t_j+k|t_j})$ is the reconstructed observation from the dynamics prediction for timestep $t_j + k$. Finally, $\hat{I}_{t_j+k} \coloneqq D_\xi(x_{t_j + k})$ is the encoded-decoded observation at timestep $t_j + k$.

Note that we use a multiple-shooting error for the loss \eqref{eq:dynamics-loss}, as the prediction horizon $T_{\text{pred}}$ does not need to coincide with the length of the trajectories in the dataset $\mathbb{D}$. In particular, the loss \eqref{eq:dynamics-loss} is computed by sampling a batch of trajectories from $\mathbb{D}$ and then splitting them into $N_{\text{dyn}}$ portions of length $T_{\text{pred}}$. The initial timestep of each portion $j = 1, ..., N_{\text{dyn}}$ is denoted as $t_j$. The first two terms in the loss function are then penalizing the state and reconstruction error of the multistep predictions of the dynamics model from each initial state $x_{t_j}$. The last term in the loss function penalizes the reconstruction error of the autoencoder directly, without using the dynamics model.
The recent results of \citet{beintema2021nonlinear} show that multiple shooting loss functions lead to more accurate predictions compared to single-step prediction losses, and to better conditioned learning problems compared to single-shooting propagation losses.

An iDBF can be learned together with the autoencoder and dynamics model by
optimizing jointly the losses \eqref{eq:cbf-loss} and \eqref{eq:dynamics-loss}. For the iDBF loss, each $\xsafe$ is obtained by encoding the observations sampled from the dataset $\mathbb{D}$, and $\xunsafe$ is obtained by forward propagating the actions that have a low probability according to the pretrained BC model, as explained at the end of last section.

Once the iDBF $B_\phi$; dynamics model $f_\theta$ and $g_\theta$; and encoder $E_\psi$ networks are trained, we can construct a policy filter ---which we call iDBF-QP in Figure \ref{fig:diagram}--- in an equivalent manner to the CBF-QP that was introduced in Section \ref{sec:02background}.
\vspace{-0.3em}
\begin{remark}
It is important to note that our iDBF training procedure encourages the satisfaction of the CBF conditions \eqref{eq:cbf-def-cond-safe}, \eqref{eq:cbf-def-cond-unsafe} and \eqref{eq:cbf-def-cond-ascent} only at a discrete set of
training points (which has measure zero). Because of this, we do not have control invariance guarantees for any particular set, and solving the iDBF-QP does not theoretically assure that the system will remain in-distribution. Although obtaining rigorous theoretical guarantees should be a priority for future work, the empirical results of Section \ref{sec:06examples} show that our framework takes a promising first step towards building effective policy filters from raw high-dimensional observations.
\end{remark}

\vspace{-1em}
\section{Examples}
\label{sec:06examples}

\begin{table}[t]
\vspace{-1em}
\addtolength{\leftskip}{-0.2cm}
\resizebox{1.02\textwidth}{!}{
\setlength\tabcolsep{1.5pt}
\begin{tabular}{|ll !{\vrule width 1.5pt} c !{\vrule width 1.5pt} c !{\vrule width 1.5pt} ccc !{\vrule width 1.5pt} ccc|}
\hline
\multicolumn{2}{|l !{\vrule width 1.5pt}}{} &
   &
   &
  \multicolumn{3}{c !{\vrule width 1.5pt}}{\textbf{BC Filter}} &
  \multicolumn{3}{c|}{\textbf{Ensemble Filter}} \\ \cline{5-10} 
\multicolumn{2}{|l!{\vrule width 1.5pt}}{\multirow{-2}{*}{\textbf{}}} &
  \multirow{-2}{*}{$\boldsymbol{\pi}_{\textbf{ref}}$} &
  \multirow{-2}{*}{\textbf{Ours}} &
  \multicolumn{1}{c|}{$p_{\text{low}}$} &
  \multicolumn{1}{c|}{$p_{\text{mid}}$} &
  $p_{\text{high}}$ &
  \multicolumn{1}{c|}{$\delta_{\text{low}}$} &
  \multicolumn{1}{c|}{$\delta_{\text{mid}}$} &
  $\delta_{\text{high}}$ \\ \noalign{\hrule height 1.5pt}
\multicolumn{1}{|l|}{} &
  \begin{tabular}[c]{@{}l@{}} \small Collision\\ \small Rate (\%)\end{tabular} &
  \cellcolor[HTML]{C0C0C0}$\mathbf{46.72\pm 8.36}$ &
  \cellcolor[HTML]{C0C0C0}\ul{$\mathbf{0.28\pm 0.27}$} &
  \multicolumn{1}{c|}{\cellcolor[HTML]{C0C0C0}$\mathbf{35.60\pm 7.20}$} &
  \multicolumn{1}{c|}{\cellcolor[HTML]{C0C0C0}$\mathbf{13.86 \pm 4.96}$} &
  \cellcolor[HTML]{C0C0C0}$\mathbf{2.48 \pm 1.57}$ &
  \multicolumn{1}{c|}{\cellcolor[HTML]{C0C0C0}$\mathbf{43.92 \pm 7.90}$} &
  \multicolumn{1}{c|}{\cellcolor[HTML]{C0C0C0}$\mathbf{43.82 \pm 7.41}$} &
  \cellcolor[HTML]{C0C0C0}$\mathbf{42.88 \pm 7.41}$ \\ \cline{2-10} 
\multicolumn{1}{|l|}{\multirow{-2}{*}{\textbf{\begin{tabular}[c]{@{}l@{}} \small Top-Down \\ \small Navigation\end{tabular}}}} &
  \begin{tabular}[c]{@{}l@{}} \small Cumulative\\ \small Intervention\end{tabular} &
  $0.0\pm 0.0$ &
  $109.2\pm 20.1$ &
  \multicolumn{1}{c|}{$85.6 \pm 6.8$} &
  \multicolumn{1}{c|}{$146.4 \pm 9.6$} &
  $189.1 \pm 11.6$ &
  \multicolumn{1}{c|}{$150.2 \pm 19.3$} &
  \multicolumn{1}{c|}{$94.5 \pm 16.3$} &
  $52.7 \pm 11.5$ \\ \noalign{\hrule height 1.5pt}
\multicolumn{1}{|l|}{} &
  \begin{tabular}[c]{@{}l@{}} \small Collision\\ \small Rate (\%)\end{tabular} &
  \cellcolor[HTML]{C0C0C0}$\mathbf{81.00\pm 0.23}$ &
  \cellcolor[HTML]{C0C0C0}\ul{$\mathbf{1.56\pm 1.20}$} &
  \multicolumn{1}{c|}{\cellcolor[HTML]{C0C0C0}$\mathbf{21.94 \pm 1.85}$} &
  \multicolumn{1}{c|}{\cellcolor[HTML]{C0C0C0}$\mathbf{14.44 \pm 2.69}$} &
  \cellcolor[HTML]{C0C0C0}$\mathbf{8.78 \pm 1.86}$ &
  \multicolumn{1}{c|}{\cellcolor[HTML]{C0C0C0}$\mathbf{78.74 \pm 0.20}$} &
  \multicolumn{1}{c|}{\cellcolor[HTML]{C0C0C0}$\mathbf{78.60 \pm 1.63}$} &
  \cellcolor[HTML]{C0C0C0}$\mathbf{81.50 \pm 0.27}$ \\ \cline{2-10} 
\multicolumn{1}{|l|}{\multirow{-2}{*}{\textbf{\begin{tabular}[c]{@{}l@{}} \small Egocentric \\ \small Driving\end{tabular}}}} &
  \begin{tabular}[c]{@{}l@{}} \small Cumulative\\ \small Intervention\end{tabular} &
  $0.0\pm 0.0$ &
  $278.1 \pm 32.6$ &
  \multicolumn{1}{c|}{$713.8 \pm 1.4$} &
  \multicolumn{1}{c|}{$726.7 \pm 2.6$} &
  $750.8 \pm 6.9$ &
  \multicolumn{1}{c|}{$28.7 \pm 2.1$} &
  \multicolumn{1}{c|}{$42.8 \pm 3.6$} &
  $208.9 \pm 5.7$ \\ \hline
\end{tabular}}
  \caption{Evaluation of the collision rate and cumulative filter intervention (a measure of how intrusive the filter is with respect to the reference controller) for the top-down view robotic navigation example (over $20$ simulations of $5$-seconds each with random initial and goal states) and for the egocentric view autonomous driving example (over $20$ simulations of $50$-seconds each with random initial heading angles). For the BC and ensemble filters, we provide results for $3$ different threshold values: $(p_{\text{low}},p_{\text{mid}}, p_{\text{high}})  = (0.32, 0.35, 0.38)$ for the navigation example, and $(0.2, 0.5, 0.8)$ for driving; and $(\delta_{\text{low}},\delta_{\text{mid}}, \delta_{\text{high}})  = (0.0005, 0.001, 0.002)$ for both examples.}
  \label{table:results}
  \vspace{-0.5em}
\end{table}

In this section, we present the empirical evaluation of our framework on two different simulation environments: a toy example of a robot navigation task using top-down images of the scene, and an autonomous driving scenario with egocentric image observations. For both cases, given a safety-agnostic reference controller $\pi_{\text{ref}}$, we use our iDBF-QP at each timestep with the latest image measurement to find the closest control input to $\pi_{\text{ref}}$ among those that prevent the system from entering OOD states (see Figure \ref{fig:diagram}). For each environment, we train the iDBF, autoencoder and dynamics model using a dataset containing $64\times 64$ RGB images of offline-collected trajectories.

\begin{figure}
    \centering
    \vspace{-1em}
    \includegraphics[width=0.9\textwidth,trim={3.0cm 1.0cm 0.5cm 1.8cm},clip]{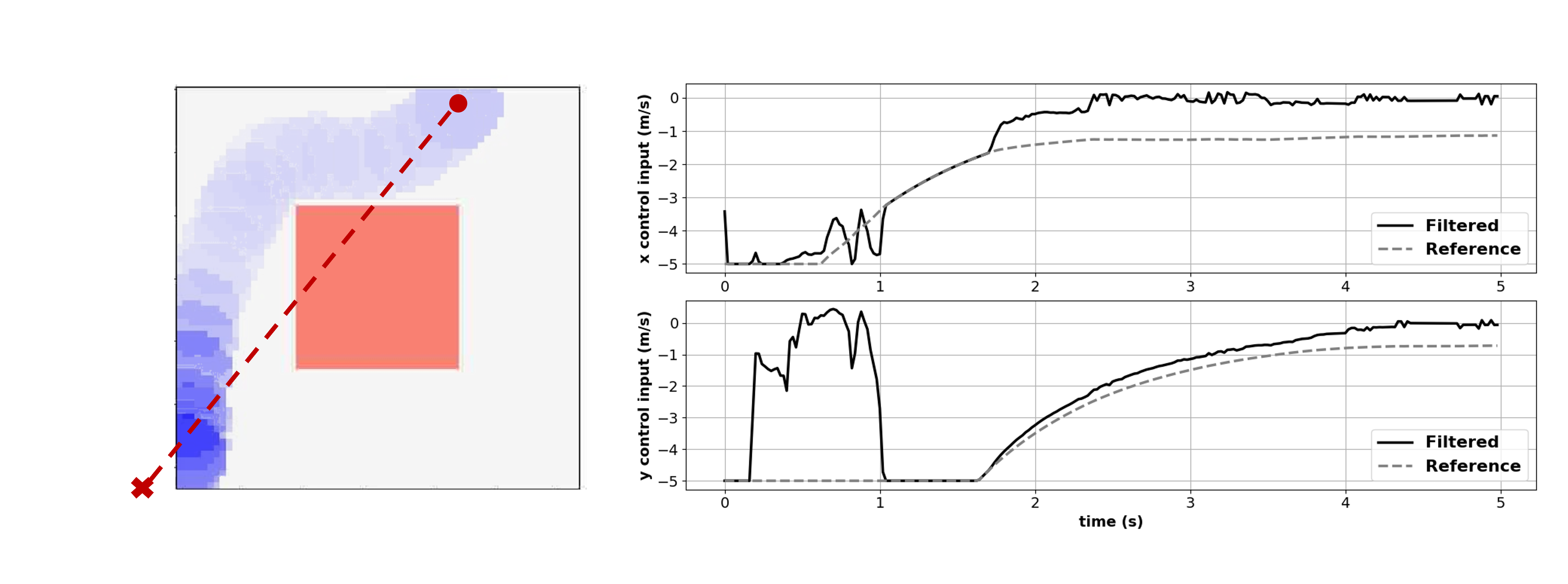}
    \caption{Example result using our proposed policy filter for a robot top-down visual navigation task. The reference controller simply tries to bring the robot (blue circle) to a goal state (denoted with $\times$). Our proposed filter, by keeping the system in-distribution, prevents the robot from colliding against the obstacle (orange square) and keeps its center-point inside the limits of the image. A video with several demonstrations of our approach for this task can be found in this \href{https://drive.google.com/drive/folders/1r57TJbgaPkdy88TAGvmPXI1GAzP4tPu1?usp=share_link}{link}.}
    \vspace{-1em}
    \label{fig:results-top-down}
\end{figure}

\textbf{Robot Navigation with Top-Down View Images:} In this example, a circular robot with radius of $1$ meter navigates inside of a $10\times10$ meter room that has a square-shaped $4\times4$ meter static obstacle in the middle, as shown in Figure \ref{fig:results-top-down} (left). The underlying dynamics of the robot are those of a 2D single integrator, with two control inputs corresponding to the $x$ and $y$ velocity commands, although we do not assume having access to that knowledge. Instead, we only have a dataset of image-action pairs corresponding to $5000$ trajectories of $100$ points each (corresponding to $2$ seconds since the time-step is $0.02$s). These trajectories satisfy two requirements: 1) the robot should never collide against the obstacle, and 2) the center of the robot should never leave the room limits. The trajectories are collected applying random actions at each time-step, and we check both conditions before adding a trajectory to the dataset. We use our framework to train an autoencoder with latent state-space of dimension $3$, a dynamics model, and an iDBF. The reference policy $\pi_{\text{ref}}$ simply applies a velocity in the direction of a goal-point, with magnitude proportional to the distance. In Figure \ref{fig:results-top-down}, we show the results of applying our iDBF-QP when the goal state (marked with an $\times$) is outside of the room limits and at the other side of the obstacle. Even though the reference controller is trying to take the shortest path, which would go through the obstacle, the iDBF-QP prevents the robot from first, colliding with the obstacle, and second, from having its center exit the room limits.

\begin{figure}
    \centering
    \vspace{-2em}
    \includegraphics[width=\textwidth]{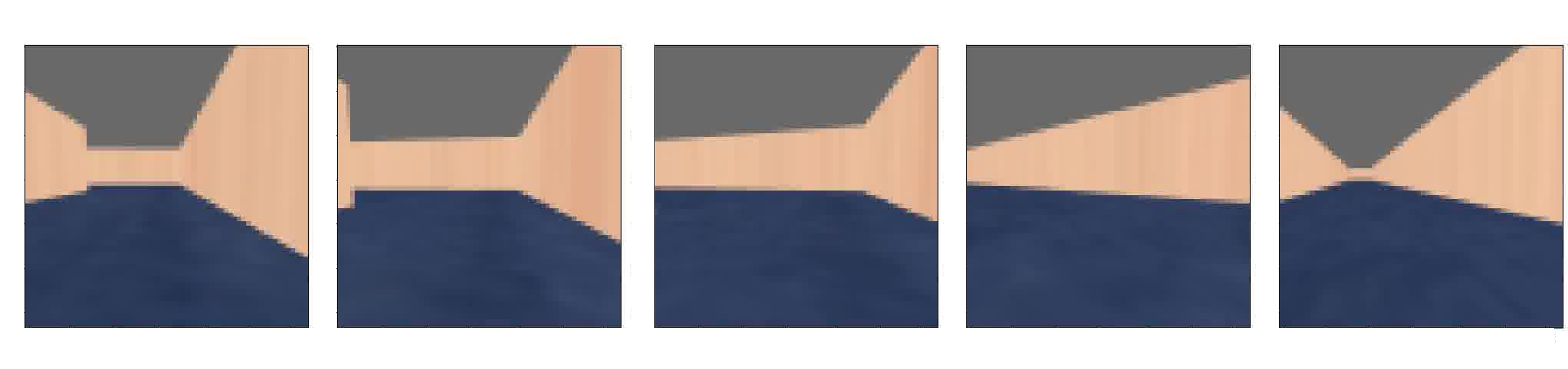}
    \vspace{-2em}
    \caption{Snapshots of egocentric view images of a driving simulation when the car is approaching a corner. The reference controller just commands the car to drive straight, but our iDBF-QP policy filter forces a left turn as the car approaches the corner. Therefore, our filter prevents a collision as a result of staying in-distribution with respect to the safe training data. A video with several demonstrations of our approach for this task can be found in this \href{https://drive.google.com/drive/folders/1r57TJbgaPkdy88TAGvmPXI1GAzP4tPu1?usp=share_link}{link}.}
    \vspace{-1em}
    \label{fig:results-driving}
\end{figure}

\textbf{Autonomous Driving with Egocentric View Images:} We use the environment provided by \citet{kahn2018self}, which is based on the Bullet physics simulator and the Panda3d graphics engine \citep{goslin2004panda3d} to obtain egocentric RGB image measurements. The car navigates in a corridor which has four 90-degree turns to form a square-shaped center-line. One of such turns is shown in the snapshots of Figure \ref{fig:results-driving}. The car has two control inputs: the desired forward velocity and the steering angle. Given the high-order dynamics of the simulator, we collect data manually to make sure no trajectories included in the dataset are deemed to collide with any of the walls. We split the collected data into $450$ trajectories of $100$ points each ($5$ seconds since the timestep is $0.05$s). This makes for a much sparser and less diverse (since it is collected by a human) dataset compared to the previous example. During deployment, we use a reference controller $\pi_{\text{ref}}$ that simply drives the car forward at a constant speed of $3.5 m/s$. Our iDBF-QP framework of Figure \ref{fig:diagram}, taking the latest egocentric RGB measurement as input, is very effective at preventing the car from colliding against the walls, as shown in Table \ref{table:results}. Figure \ref{fig:results-driving} contains snapshots of our iDBF-QP forcing the car to take a turn as it approaches a corner, even though the reference command is to drive forward.

Using these simulation environments we also aim to compare our proposed approach with other techniques for avoiding distributional shift. Other works that consider this problem use data density models to constrain the learned policies \citep{richter2017safe, mcallister2019robustness, wu2019behavior}, or use uncertainty estimation schemes, such as ensemble models, to avoid taking actions that lead to highly uncertain states \citep{chua2018deep}. We build our baselines upon a conditional BC density model of the training data and an ensemble of latent state-space dynamics models:
    
    \textbf{BC Density Filter Baseline:} As explained in Section \ref{sec:04learningcbfs}, we train a BC multi-modal Gaussian model that is used to generate the contrastive training distribution for the iDBF. For any state, the BC model outputs a probability distribution over actions, with density function $\pi_{\text{BC}} (u | x)$. We train this BC model using privileged true-state information of the system, and use its density values to build a filter that serves as an apples-to-apples baseline comparison to our approach.
    Specifically, the baseline also takes the reference controller $\pi_{\text{ref}}$ and, at every timestep, it finds the closest control action to $\pi_{\text{ref}}(x)$ that satisfies $\pi_{\text{BC}} (u | x) \geq p$, out of $200$ randomly sampled actions.
    If no control action satisfying that condition is found, the reference control input is applied without filtering. Given the clear dependence on the threshold value $p$, we implement this baseline for several values of $p$ and show the results in Table \ref{table:results} for three representative cases $p_{\text{low}},\ p_{\text{mid}}$ and $p_{\text{high}}$.
    
    \textbf{Ensemble Variance Filter Baseline:} We also train an ensemble of independent latent state-space dynamic models ($f_\theta$ and $g_\theta$), keeping the rest of the framework introduced in Section \ref{sec:05learningcbfsinlatentspace} unchanged. During deployment, at every timestep we look for the closest control action to $\pi_{\text{ref}}(x)$ that keeps the variance $\sigma^2_{\text{ens}}(x,u)$ of the predicted dynamics $f_\theta(x) + g_\theta(x) u$ under a threshold $\delta$. As in the previous baseline, we also look over $200$ randomly sampled actions at each timestep, and different threshold levels $\delta_{\text{low}}, \ \delta_{\text{mid}}$ and $\delta_{\text{high}}$. Again, if no control action satisfying the threshold condition is found, the reference control input is applied without filtering.

In Table \ref{table:results}, we provide a summary of the comparison results for both environments. We use the collision rate as a proxy for distributional shift, since the training data only includes collision-free trajectories. The collision rate for the robot navigation example is computed as the fraction of time that the robot spends either in collision with the obstacle or having its center-point outside of the room limits. For the driving scenario, the collision rate is the fraction of time that the robot is in collision with any of the walls. For both examples, our method drastically reduces the collision rate compared to using the reference (unfiltered) controller. Furthermore, we achieve the lowest collision rates when compared to the baselines. From the baselines, only the BC density filter (with a very restrictive threshold $p_{\text{high}}$) manages to achieve small collision rates, at the cost of a very high cumulative filter intervention rate. The filter intervention rate is computed for both examples as $\sum_{t} \norm{u_t - \pi_{\text{ref}}(x_t)}^2$, where each control input dimension is normalized between $-1$ and $1$.

\section{Conclusion}
\label{sec:07conclusion}
In this work, we take first-steps towards merging control-theoretic CBFs with practical robotic tasks that involve high-dimensional perception modules. We consider a realistic problem setting in which no unsafe demonstrations are available, and take a self-supervised learning approach to learn a function that effectively restricts the system from diverging towards OOD states. By learning this function in a latent state-space, our framework should be flexible-enough to be applicable to a wide variety of visuomotor tasks, and should be compatible with the use of large-scale pretrained representation learning models. Another important direction for future work would be to use probabilistic encoding and dynamics models to be able to robustify our proposed filters with respect to prediction uncertainties. Additionally, exploring the use of loss functions that are not based on reconstruction, by exploiting the value function nature of the iDBF, could be another promising direction.

\acks{The authors would like to thank Dr. Jean Mercat, Dr. Hongkai Dai and Dr. Katherine Liu for their insightful comments and suggestions.}

\bibliography{references.bib}

\begin{thebibliography}{47}
\providecommand{\natexlab}[1]{#1}
\providecommand{\url}[1]{\texttt{#1}}
\expandafter\ifx\csname urlstyle\endcsname\relax
  \providecommand{\doi}[1]{doi: #1}\else
  \providecommand{\doi}{doi: \begingroup \urlstyle{rm}\Url}\fi

\bibitem[Abate et~al.(2021)Abate, Ahmed, Edwards, Giacobbe, and
  Peruffo]{abate2021fossil}
Alessandro Abate, Daniele Ahmed, Alec Edwards, Mirco Giacobbe, and Andrea
  Peruffo.
\newblock Fossil: a software tool for the formal synthesis of lyapunov
  functions and barrier certificates using neural networks.
\newblock In \emph{Proceedings of the 24th International Conference on Hybrid
  Systems: Computation and Control}, pages 1--11, 2021.

\bibitem[{Ames} et~al.(2014){Ames}, {Grizzle}, and {Tabuada}]{ames2014cbf}
A.~D. {Ames}, J.~W. {Grizzle}, and P.~{Tabuada}.
\newblock Control barrier function based quadratic programs with application to
  adaptive cruise control.
\newblock In \emph{IEEE Conference on Decision and Control}, pages 6271--6278,
  2014.

\bibitem[{Ames} et~al.(2017){Ames}, {Xu}, {Grizzle}, and
  {Tabuada}]{ames2017cbf}
A.~D. {Ames}, X.~{Xu}, J.~W. {Grizzle}, and P.~{Tabuada}.
\newblock Control barrier function based quadratic programs for safety critical
  systems.
\newblock \emph{IEEE Transactions on Automatic Control}, 62:\penalty0
  3861--3876, 2017.

\bibitem[Beintema et~al.(2021)Beintema, Toth, and
  Schoukens]{beintema2021nonlinear}
Gerben Beintema, Roland Toth, and Maarten Schoukens.
\newblock Nonlinear state-space identification using deep encoder networks.
\newblock In \emph{Learning for Dynamics and Control}, pages 241--250. PMLR,
  2021.

\bibitem[Castañeda et~al.(2021)Castañeda, Choi, Zhang, Tomlin, and
  Sreenath]{castaneda2021pointwise}
Fernando Castañeda, Jason~J. Choi, Bike Zhang, Claire~J. Tomlin, and Koushil
  Sreenath.
\newblock Pointwise feasibility of gaussian process-based safety-critical
  control under model uncertainty.
\newblock In \emph{IEEE Conference on Decision and Control}, pages 6762--6769,
  2021.

\bibitem[Chang et~al.(2019)Chang, Roohi, and Gao]{chang2019neural}
Ya-Chien Chang, Nima Roohi, and Sicun Gao.
\newblock Neural lyapunov control.
\newblock \emph{Advances in neural information processing systems}, 32, 2019.

\bibitem[Chen et~al.(2018)Chen, Rubanova, Bettencourt, and
  Duvenaud]{chen2018neural}
Ricky~TQ Chen, Yulia Rubanova, Jesse Bettencourt, and David~K Duvenaud.
\newblock Neural ordinary differential equations.
\newblock \emph{Advances in neural information processing systems}, 31, 2018.

\bibitem[Choi et~al.(2021)Choi, Lee, Sreenath, Tomlin, and
  Herbert]{choi2021robust}
Jason~J Choi, Donggun Lee, Koushil Sreenath, Claire~J Tomlin, and Sylvia~L
  Herbert.
\newblock Robust control barrier--value functions for safety-critical control.
\newblock In \emph{2021 60th IEEE Conference on Decision and Control (CDC)},
  pages 6814--6821. IEEE, 2021.

\bibitem[Chopra et~al.(2005)Chopra, Hadsell, and LeCun]{chopra2005learning}
Sumit Chopra, Raia Hadsell, and Yann LeCun.
\newblock Learning a similarity metric discriminatively, with application to
  face verification.
\newblock In \emph{2005 IEEE Computer Society Conference on Computer Vision and
  Pattern Recognition (CVPR'05)}, volume~1, pages 539--546. IEEE, 2005.

\bibitem[Chua et~al.(2018)Chua, Calandra, McAllister, and Levine]{chua2018deep}
Kurtland Chua, Roberto Calandra, Rowan McAllister, and Sergey Levine.
\newblock Deep reinforcement learning in a handful of trials using
  probabilistic dynamics models.
\newblock \emph{Advances in neural information processing systems}, 31, 2018.

\bibitem[Cosner et~al.(2022)Cosner, Rodriguez, Molnar, Ubellacker, Yue, Ames,
  and Bouman]{cosner2022self}
Ryan~K Cosner, Ivan D~Jimenez Rodriguez, Tamas~G Molnar, Wyatt Ubellacker,
  Yisong Yue, Aaron~D Ames, and Katherine~L Bouman.
\newblock Self-supervised online learning for safety-critical control using
  stereo vision.
\newblock \emph{arXiv preprint arXiv:2203.01404}, 2022.

\bibitem[Dai and Permenter(2022)]{dai2022convex}
Hongkai Dai and Frank Permenter.
\newblock Convex synthesis and verification of control-lyapunov and barrier
  functions with input constraints.
\newblock \emph{arXiv preprint arXiv:2210.00629}, 2022.

\bibitem[Dawson et~al.(2022)Dawson, Qin, Gao, and Fan]{dawson2022safe}
Charles Dawson, Zengyi Qin, Sicun Gao, and Chuchu Fan.
\newblock Safe nonlinear control using robust neural lyapunov-barrier
  functions.
\newblock In \emph{Conference on Robot Learning}, pages 1724--1735. PMLR, 2022.

\bibitem[Dean et~al.(2021)Dean, Taylor, Cosner, Recht, and
  Ames]{dean2021guaranteeing}
Sarah Dean, Andrew Taylor, Ryan Cosner, Benjamin Recht, and Aaron Ames.
\newblock Guaranteeing safety of learned perception modules via
  measurement-robust control barrier functions.
\newblock In \emph{Conference on Robot Learning}, 2021.

\bibitem[Dhiman et~al.(2021)Dhiman, Khojasteh, Franceschetti, and
  Atanasov]{dhiman2021control}
Vikas Dhiman, Mohammad~Javad Khojasteh, Massimo Franceschetti, and Nikolay
  Atanasov.
\newblock Control barriers in bayesian learning of system dynamics.
\newblock \emph{IEEE Transactions on Automatic Control}, 2021.

\bibitem[Ebert et~al.(2018)Ebert, Finn, Dasari, Xie, Lee, and
  Levine]{ebert2018visual}
Frederik Ebert, Chelsea Finn, Sudeep Dasari, Annie Xie, Alex Lee, and Sergey
  Levine.
\newblock Visual foresight: Model-based deep reinforcement learning for
  vision-based robotic control.
\newblock \emph{arXiv preprint arXiv:1812.00568}, 2018.

\bibitem[Goslin and Mine(2004)]{goslin2004panda3d}
Mike Goslin and Mark~R Mine.
\newblock The panda3d graphics engine.
\newblock \emph{Computer}, 37\penalty0 (10):\penalty0 112--114, 2004.

\bibitem[Gutmann and Hyv{\"a}rinen(2010)]{gutmann2010noise}
Michael Gutmann and Aapo Hyv{\"a}rinen.
\newblock Noise-contrastive estimation: A new estimation principle for
  unnormalized statistical models.
\newblock In \emph{Proceedings of the thirteenth international conference on
  artificial intelligence and statistics}, pages 297--304. JMLR Workshop and
  Conference Proceedings, 2010.

\bibitem[Gutmann and Hirayama(2011)]{gutmann2011bregman}
Michael~U Gutmann and Jun-ichiro Hirayama.
\newblock Bregman divergence as general framework to estimate unnormalized
  statistical models.
\newblock In \emph{Proceedings of the Twenty-Seventh Conference on Uncertainty
  in Artificial Intelligence}, pages 283--290, 2011.

\bibitem[Hafner et~al.(2019)Hafner, Lillicrap, Fischer, Villegas, Ha, Lee, and
  Davidson]{hafner2019learning}
Danijar Hafner, Timothy Lillicrap, Ian Fischer, Ruben Villegas, David Ha,
  Honglak Lee, and James Davidson.
\newblock Learning latent dynamics for planning from pixels.
\newblock In \emph{International conference on machine learning}, pages
  2555--2565. PMLR, 2019.

\bibitem[Hinton(2002)]{hinton2002training}
Geoffrey~E Hinton.
\newblock Training products of experts by minimizing contrastive divergence.
\newblock \emph{Neural computation}, 14\penalty0 (8):\penalty0 1771--1800,
  2002.

\bibitem[Jagtap et~al.(2020)Jagtap, Pappas, and Zamani]{jagtap2020control}
Pushpak Jagtap, George~J Pappas, and Majid Zamani.
\newblock Control barrier functions for unknown nonlinear systems using
  gaussian processes.
\newblock In \emph{2020 59th IEEE Conference on Decision and Control (CDC)},
  pages 3699--3704. IEEE, 2020.

\bibitem[Jarvis-Wloszek et~al.(2003)Jarvis-Wloszek, Feeley, Tan, Sun, and
  Packard]{jarvis2003some}
Zachary Jarvis-Wloszek, Ryan Feeley, Weehong Tan, Kunpeng Sun, and Andrew
  Packard.
\newblock Some controls applications of sum of squares programming.
\newblock In \emph{42nd IEEE international conference on decision and control},
  volume~5, pages 4676--4681. IEEE, 2003.

\bibitem[Jin et~al.(2020)Jin, Wang, Yang, and Mou]{jin2020neural}
Wanxin Jin, Zhaoran Wang, Zhuoran Yang, and Shaoshuai Mou.
\newblock Neural certificates for safe control policies.
\newblock \emph{arXiv preprint arXiv:2006.08465}, 2020.

\bibitem[Kahn et~al.(2018)Kahn, Villaflor, Ding, Abbeel, and
  Levine]{kahn2018self}
Gregory Kahn, Adam Villaflor, Bosen Ding, Pieter Abbeel, and Sergey Levine.
\newblock Self-supervised deep reinforcement learning with generalized
  computation graphs for robot navigation.
\newblock In \emph{2018 IEEE International Conference on Robotics and
  Automation (ICRA)}, pages 5129--5136. IEEE, 2018.

\bibitem[Kang et~al.(2022)Kang, Gradu, Choi, Janner, Tomlin, and
  Levine]{kang2022lyapunov}
Katie Kang, Paula Gradu, Jason~J Choi, Michael Janner, Claire Tomlin, and
  Sergey Levine.
\newblock Lyapunov density models: Constraining distribution shift in
  learning-based control.
\newblock In \emph{International Conference on Machine Learning}, pages
  10708--10733. PMLR, 2022.

\bibitem[Kumar et~al.(2019)Kumar, Fu, Soh, Tucker, and
  Levine]{kumar2019stabilizing}
Aviral Kumar, Justin Fu, Matthew Soh, George Tucker, and Sergey Levine.
\newblock Stabilizing off-policy q-learning via bootstrapping error reduction.
\newblock \emph{Advances in Neural Information Processing Systems}, 32, 2019.

\bibitem[Lenz et~al.(2015)Lenz, Knepper, and Saxena]{lenz2015deepmpc}
Ian Lenz, Ross~A Knepper, and Ashutosh Saxena.
\newblock Deepmpc: Learning deep latent features for model predictive control.
\newblock In \emph{Robotics: Science and Systems}, volume~10. Rome, Italy,
  2015.

\bibitem[Lindemann et~al.(2021)Lindemann, Robey, Jiang, Tu, and
  Matni]{lindemann2021learning}
Lars Lindemann, Alexander Robey, Lejun Jiang, Stephen Tu, and Nikolai Matni.
\newblock Learning robust output control barrier functions from safe expert
  demonstrations.
\newblock \emph{arXiv preprint arXiv:2111.09971}, 2021.

\bibitem[Long et~al.(2021)Long, Qian, Cort{\'e}s, and
  Atanasov]{long2021learning}
Kehan Long, Cheng Qian, Jorge Cort{\'e}s, and Nikolay Atanasov.
\newblock Learning barrier functions with memory for robust safe navigation.
\newblock \emph{IEEE Robotics and Automation Letters}, 6\penalty0 (3):\penalty0
  4931--4938, 2021.

\bibitem[Majumdar et~al.(2013)Majumdar, Ahmadi, and
  Tedrake]{majumdar2013control}
Anirudha Majumdar, Amir~Ali Ahmadi, and Russ Tedrake.
\newblock Control design along trajectories with sums of squares programming.
\newblock In \emph{2013 IEEE International Conference on Robotics and
  Automation}, pages 4054--4061. IEEE, 2013.

\bibitem[McAllister et~al.(2019)McAllister, Kahn, Clune, and
  Levine]{mcallister2019robustness}
Rowan McAllister, Gregory Kahn, Jeff Clune, and Sergey Levine.
\newblock Robustness to out-of-distribution inputs via task-aware generative
  uncertainty.
\newblock In \emph{2019 International Conference on Robotics and Automation
  (ICRA)}, pages 2083--2089. IEEE, 2019.

\bibitem[{Nguyen} and {Sreenath}(2021)]{nguyen2021robust}
Q.~{Nguyen} and K.~{Sreenath}.
\newblock Robust safety-critical control for dynamic robotics.
\newblock \emph{IEEE Transactions on Automatic Control}, 2021.

\bibitem[Oord et~al.(2018)Oord, Li, and Vinyals]{oord2018representation}
Aaron van~den Oord, Yazhe Li, and Oriol Vinyals.
\newblock Representation learning with contrastive predictive coding.
\newblock \emph{arXiv preprint arXiv:1807.03748}, 2018.

\bibitem[Qin et~al.(2022)Qin, Sun, and Fan]{qin2022sablas}
Zengyi Qin, Dawei Sun, and Chuchu Fan.
\newblock Sablas: Learning safe control for black-box dynamical systems.
\newblock \emph{IEEE Robotics and Automation Letters}, 7\penalty0 (2):\penalty0
  1928--1935, 2022.

\bibitem[Richter and Roy(2017)]{richter2017safe}
Charles Richter and Nicholas Roy.
\newblock Safe visual navigation via deep learning and novelty detection.
\newblock In \emph{Robotics: Science and Systems}. Cambridge, MA, 2017.

\bibitem[Robey et~al.(2020)Robey, Hu, Lindemann, Zhang, Dimarogonas, Tu, and
  Matni]{robey2020learning}
Alexander Robey, Haimin Hu, Lars Lindemann, Hanwen Zhang, Dimos~V Dimarogonas,
  Stephen Tu, and Nikolai Matni.
\newblock Learning control barrier functions from expert demonstrations.
\newblock In \emph{2020 59th IEEE Conference on Decision and Control (CDC)},
  pages 3717--3724. IEEE, 2020.

\bibitem[Schroff et~al.(2015)Schroff, Kalenichenko, and
  Philbin]{schroff2015facenet}
Florian Schroff, Dmitry Kalenichenko, and James Philbin.
\newblock Facenet: A unified embedding for face recognition and clustering.
\newblock In \emph{Proceedings of the IEEE conference on computer vision and
  pattern recognition}, pages 815--823, 2015.

\bibitem[Srinivasan et~al.(2020)Srinivasan, Dabholkar, Coogan, and
  Vela]{srinivasan2020synthesis}
Mohit Srinivasan, Amogh Dabholkar, Samuel Coogan, and Patricio~A Vela.
\newblock Synthesis of control barrier functions using a supervised machine
  learning approach.
\newblock In \emph{2020 IEEE/RSJ International Conference on Intelligent Robots
  and Systems (IROS)}, pages 7139--7145. IEEE, 2020.

\bibitem[Taylor et~al.(2021)Taylor, Dorobantu, Dean, Recht, Yue, and
  Ames]{taylor2020towards}
Andrew~J. Taylor, Victor~D. Dorobantu, Sarah Dean, Benjamin Recht, Yisong Yue,
  and Aaron~D. Ames.
\newblock Towards robust data-driven control synthesis for nonlinear systems
  with actuation uncertainty.
\newblock In \emph{IEEE Conference on Decision and Control}, pages 6469--6476,
  2021.

\bibitem[Van~Hoof et~al.(2016)Van~Hoof, Chen, Karl, van~der Smagt, and
  Peters]{van2016stable}
Herke Van~Hoof, Nutan Chen, Maximilian Karl, Patrick van~der Smagt, and Jan
  Peters.
\newblock Stable reinforcement learning with autoencoders for tactile and
  visual data.
\newblock In \emph{2016 IEEE/RSJ international conference on intelligent robots
  and systems (IROS)}, pages 3928--3934. IEEE, 2016.

\bibitem[Wang et~al.(2022)Wang, Margellos, and
  Papachristodoulou]{wang2022safety}
Han Wang, Kostas Margellos, and Antonis Papachristodoulou.
\newblock Safety verification and controller synthesis for systems with input
  constraints.
\newblock \emph{arXiv preprint arXiv:2204.09386}, 2022.

\bibitem[Watter et~al.(2015)Watter, Springenberg, Boedecker, and
  Riedmiller]{watter2015embed}
Manuel Watter, Jost Springenberg, Joschka Boedecker, and Martin Riedmiller.
\newblock Embed to control: A locally linear latent dynamics model for control
  from raw images.
\newblock \emph{Advances in neural information processing systems}, 28, 2015.

\bibitem[Weinberger and Saul(2009)]{weinberger2009distance}
Kilian~Q Weinberger and Lawrence~K Saul.
\newblock Distance metric learning for large margin nearest neighbor
  classification.
\newblock \emph{Journal of machine learning research}, 10\penalty0 (2), 2009.

\bibitem[Wilcox et~al.(2022)Wilcox, Balakrishna, Thananjeyan, Gonzalez, and
  Goldberg]{wilcox2022ls3}
Albert Wilcox, Ashwin Balakrishna, Brijen Thananjeyan, Joseph~E Gonzalez, and
  Ken Goldberg.
\newblock Ls3: Latent space safe sets for long-horizon visuomotor control of
  sparse reward iterative tasks.
\newblock In \emph{Conference on Robot Learning}, pages 959--969. PMLR, 2022.

\bibitem[Wu et~al.(2019)Wu, Tucker, and Nachum]{wu2019behavior}
Yifan Wu, George Tucker, and Ofir Nachum.
\newblock Behavior regularized offline reinforcement learning.
\newblock \emph{arXiv preprint arXiv:1911.11361}, 2019.

\bibitem[Zhang et~al.(2019)Zhang, Vikram, Smith, Abbeel, Johnson, and
  Levine]{zhang2019solar}
Marvin Zhang, Sharad Vikram, Laura Smith, Pieter Abbeel, Matthew Johnson, and
  Sergey Levine.
\newblock Solar: Deep structured representations for model-based reinforcement
  learning.
\newblock In \emph{International Conference on Machine Learning}, pages
  7444--7453. PMLR, 2019.

\end{thebibliography}

\end{document}